# Adaptive search area for fast motion estimation

S.M.Reza Soroushmehr, Shadrokh Samavi, Shahram Shirani

**Abstract**: This paper suggests a new method for determining the search area for a motion estimation algorithm based on block matching. The search area is adaptively found in the proposed method for each frame block. This search area is similar to that of the full search (FS) algorithm but smaller for most blocks of a frame. Therefore, the proposed algorithm is analogous to FS in terms of regularity but has much less computational complexity. The temporal and spatial correlations among the motion vectors of blocks are used to find the search area. The matched block is chosen from a rectangular area that the prediction vectors set out. Simulation results indicate that the speed of the proposed algorithm is at least seven times better than the FS algorithm.

## 1 Introduction

Motion estimation has a key role in compression of video sequences. Using motion estimation one can reduce temporal correlation among consecutive frames and hence reduce the data volume that is needed to store or transmit the sequences. During the past two decades many algorithms have been offered for motion estimation. Algorithms such as Pel Recursive Algorithm (PRA) [1], Transform domain [2], Gradient techniques, algorithms that use a mesh and Block Matching Algorithms (BMA) are among motion estimation algorithms. Due to simplicity and high performance BMA is used in video coding and compression standards such as *MPEG1/2/4, H.261, H.263,* and *H.264.*
Full Search (FS) algorithm is a block matching routine. This algorithm is simple and regular and hence hardware implementations usually use it [4]. Another advantage of the FS beside regularity is its ability to find global minimum point. Use of FS algorithm causes difficulty motion estimation implementation and is responsible for 75% of the coder's computational complexity. Applications such as video phones, video conferencing and video recording require fast methods for image coding. High compression ratio and generation of high quality reconstructed images are among other characteristics of an ideal algorithm. Many algorithms have been devised to satisfy a part of the above mentioned qualities.

In this paper we offer a block matching algorithm which uses the temporal and spatial correlation among motion vectors. The suggested method locates a search area for each individual block. This is rectangular area that spans over all neighboring motion vectors.

Regularity of the method is similar to FS but depending on the type of image and quality of the reconstructed frames the search area can be much smaller than that of FS.

The rest of the paper is organized in the following manner: In section II block matching motion estimation algorithms are reviewed. Due to its importance in our investigation, section III is dedicated to the review of the predictive search algorithm (PSA) [11]. Our suggested method of determining the search-area along with the proposed algorithm are explained in section IV. Simulation results are presented in section V. Concluding remarks are offered in the final section.

## 2 Block matching motion estimation

In motion estimation algorithms that are based on block matching, a frame is divided into a number of non-overlapped $N*N$ blocks. Then for each block a search area in the reference frame is designated. This area is the outward extension of W pixels from boundaries of the block in the reference frame. In order to find the matched block a criterion function is required. Criterions such as mean square error (MSE), sum square error (SSE), mean absolute error (MAE) and sum absolute error (SAE) are defined by Equation (1) by setting $(\beta, \delta)$ respectively equal to (1,2), (0,2), (1,1), and (0,1).

$$E(x, y) = (1/N^2)^\beta \times \sum_{m=0}^{N-1}\sum_{n=0}^{N-1} \left| I_{cur}(x_0+m, y_0+n) - I_{ref}(x_0+x+m, y+y_0+n) \right|^\delta \quad (1)$$

where $-W \leq (x, y) \leq W$

Due to simple implementation SAE and MAE are used more often. In this equation $I_{ref}(i,j)$ and $I_{cur}(i,j)$ are intensities of pixels at coordinates (i, j) in the reference and current frames.

Within the search area the block that minimizes the above criterion is chosen as the matched block. The vector that connects a point on the current block to a corresponding point on the matched block is called motion vector. Figure 1 shows an example of a search area and a motion vector.

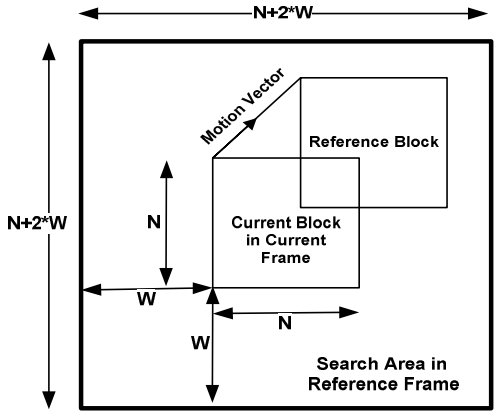

**Fig 1**. Illustration of search area and motion vector.

In the FS algorithm all of the $(2w+1)^2$ blocks of the search area have to be tested. While being simple and regular, FS requires high computational efforts. To reduce computational complexity a number of other search algorithms have been suggested. In some of the algorithms there are fixed search patterns for finding the best matched block. Rood pattern algorithm (RPA) [1], logarithmic search algorithm [5], three step search (3SS) [6], four step search (4SS) [7], diamond search (DS) [8], and hexagonal search pattern [9] use fixed search area. In all of these algorithms unimodal error surface assumption (UESA) is considered [5]. This assumption is not always true. Therefore, this group of algorithms may be trapped in a local minimum. Of course this UESA can be true in a small area around the global minimum point [10]. Hence, in some routines such as predictive search algorithm initially the motion vector is predicted and then the searching is performed around the predicted vector [11]. This algorithm searches for different regions to find the minimum point. These regions may be in any location inside the general search area. The regions may have overlaps or they may have no common points. Therefore, in general, there is no regularity in the search pattern. In the next section we will review the PSA's method for determining these search regions.

Different hardware schemes have been devised for real-time motion estimation. Most of the implementations have used FS because of its regularity. Algorithms such as 3SS and 4SS have also been implemented but due to low degree of regularity their VLSI realizations have not been attractive [4]. In 3SS at each step 9 points are searched. The center of the search point is the minimum point of the last step. To find the center of the search pattern at each step 9 points have to be tested. Also, the distance between the search points at one step is different than those of other steps.

Therefore, we use the same pattern as the FS algorithm in order to make the hardware implementation more efficient.

## 3 Predictive search algorithm

Due to the importance of the predictive search algorithm [11] in our investigation we explain this algorithm in some details in this section.

In PSA algorithm, according to the observations on the statistical distribution of motion differentials among the motion vectors of any block and those of its four neighboring blocks from six real video sequences, a new predictive search area approach is suggested by Chung. Before explaining the observations, a number of definitions are required.

Variable D is defined in Equation (2).
$$D = \min\{\max\{|MVXc - MVXi|, |MVYc - MVYi|\}\}$$
$$(1 \leq i \leq 4) \quad (2)$$

where D is the displacement of the motion vector differentials between the Bc block and the nearest neighboring block in pixel.

Here, $MVXi$, $MVYi$ are respectively the displacement of block Bi in the horizontal and vertical directions with respect to the original block. Also $MVXc$, $MVYc$ are respectively the motion vector of current block (Bc). These blocks are displayed in figure 2.

For the $J^{th}$ image frame in the video sequence L ($1 \leq L \leq 6$), let $Pr_{J,L}(D = d)$ denote the probability when D = d. The size of the search window used in PSA experiment is 33*33, so d must be between 1 and 16. Except the first image frame in the $L^{th}$ video sequence, for the remaining image frames in the same video sequence, the average probability of D = d is defined by

$$Prob_L(D=d) = \frac{1}{n_L - 1}\sum_{J=2}^{n_L} Pr_{J,L}(D=d) \quad (3)$$

where $n_L$ denotes the number of image frames in the video sequence L. The average probability is defined by

$$Prob_{n_L}(D=d) = 1/6 \sum_{L=0}^{5} Prob_L \quad (4)$$

the accumulated probability is defined by
$$Prob_{n_L}(D \leq d) =$$
$$1/6 \sum_{L=0}^{5} \frac{1}{n_L - 1} \sum_{J=2}^{n_L} \sum_{D=0}^{d} Pr_{J,L}(D=d) \quad (5)$$

According to Chung's simulations, he observed that the average accumulated probability is 94.24% for D = 2 and 95.54% for D = 3.

From this observation, due to the high probability, (i.e. 94.24%), D = 2 is a good choice to confine the search area for the current block to find the best matching block in the reference image frame. Of course, D = 3 is also an applicable choice to confine the search area for the current block Bc.

Based on these results, PSA places a 2x2 search area around each of the four prediction vectors. Hence,

irrespective of the possible overlapped point, 100 point need to be checked for each block. In this paper a new algorithm is proposed that has similarities with PSA but has many advantages to that due to difference in terms of structure and performance. In the next section we explain the suggested algorithm.

## 4 Proposed algorithm

In this section we explain the details of our algorithm and introduce how the search area is determined for each block. In order to find the boundaries of the search area, we first performed some statistical analyses by applying the FS algorithm to eight standard video sequences with different video characteristics. In these analyses the occasions that the motion vector of a block falls within a specific rectangular area is computed. The mentioned rectangular area is determined by the prediction vectors that are the motion vectors of the neighboring blocks.

Since an object usually occupies more than one block of an image, motions of neighboring blocks are similar to each other. This is known as spatial

$$P(d)_{(m,n,f)} = \Pr((DiffMinX_{(m,n,f)} \le d) \cap (DiffMinY_{(m,n,f)} \le d) \cap (DiffMaxX_{(m,n,f)} \le d) \cap (DiffMaxY_{(m,n,f)} \le d)) \quad (10)$$

correlation of motion vectors. Also, due to inertia in the movement of objects, there is correlation among motion vectors of blocks of consecutive frames. This is known as temporal correlation of motion vectors. It has also been shown that the directions that the neighboring blocks move are similar [12, 13]. It is hence expected that the motion vectors of neighboring blocks fall within a small region. This is proved by simulations.

In Figure 2 a number of blocks of a frame are shown. In the FS algorithm the motion vector of blocks are found in a row by row manner starting from top left corner. Therefore, in Figure 2 when we get to the block that is called Bc all the blocks that have a check mark,✓, have known motion vector.

In this work we use the motion vector of blocks *B1* to *B4* because of their higher spatial correlation with *Bc* as compared to other close by blocks. Also, we use the motion vector of the block corresponding to *Bc* in the reference frame. We refer to this block as *B5*.

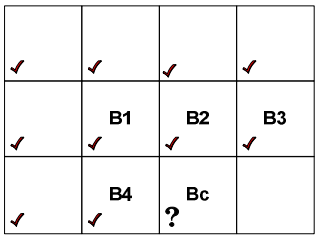

**Fig. 2**. Block Bc and its neighboring blocks.

We show motion vectors of blocks Bi as $(MVXi, MVYi)$ where $(1 \le i \le 5)$.

To come up with the statistical analysis we define the following parameters:

$$DiffMinX_{(m,n,f)} = \min_{1 \le i \le 5}(MVXi) - MVXc_{(m,n,f)} \quad (6)$$

$$DiffMaxX_{(m,n,f)} = MVXc_{(m,n,f)} - \max_{1 \le i \le 5}(MVXi) \quad (7)$$

$$DiffMinY_{(m,n,f)} = \min_{1 \le i \le 5}(MVYi) - MVYc_{(m,n,f)} \quad (8)$$

$$DiffMaxY_{(m,n,f)} = MVYc_{(m,n,f)} - \max_{1 \le i \le 5}(MVYi) \quad (9)$$

In the above expressions $DiffMinX_{(m,n,f)}$ and $DiffMinY_{(m,n,f)}$ are the differences between displacement of a block in a frame f and the *minimum* displacement among the neighboring blocks respectively in the horizontal and vertical directions. With the same token, $DiffMaxX_{(m,n,f)}$ and $DiffMaxY_{(m,n,f)}$ are the differences between displacement of a block in a frame f and the *maximum* displacement among the neighboring blocks respectively in the horizontal and vertical directions. It is assumed that the coordinates of the upper left of the block is (m,n).

Equation (10) defines $P(d)_{(m,n,f)}$ which is the probability that the motion vector of a block with the coordinates (m,n) in a frame, f, falls in a rectangular area with side lengths of $\max_{1 \le i \le 5}(MVX_i) - \min_{1 \le i \le 5}(MVX_i) + 2d$ and $\max_{1 \le i \le 5}(MVY_i) - \min_{1 \le i \le 5}(MVY_i) + 2d$.

Figure 3 shows the mentioned rectangular area for d=2. In this Figure the motion vectors of blocks B1 to B5 are respectively (3, 7), (1, 6), (-1, 5), (0, 6), and (3, 5).

Therefore, minimum and maximum values in the two directions are $\min_{1 \le i \le 5}(MVY_i) = 5$, $\min_{1 \le i \le 5}(MVX_i) = -1$, $\max_{1 \le i \le 5}(MVY_i) = 7$, and $\max_{1 \le i \le 5}(MVX_i) = 3$. Extending these values by two points (d=2) generates a rectangle with the vertices at (-3, 3), (-3, 9), (5, 3), and (5, 9).

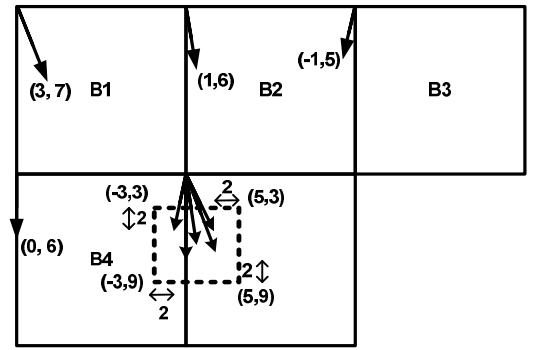

**Fig. 3.** Illustration of rectangular search area.

Assuming NF as the total number of frames in a video sequence then we define a probability Pr(*d*) as in Equation (11).

$$\Pr(d) = \frac{\sum_{f=0}^{NF-1} \sum_{m=0}^{NR-1} \sum_{n=0}^{NC-1} P(d)_{(m,n,f)}}{(NF \times NR \times NC)} \quad (11)$$

Where NR and NC are respectively the number of blocks in a row and a column of a frame. Table (1) presents the value of $\Pr(d)$ for 8 different video sequences. For example, in Susie sequence if we generate the rectangular area with d equal to 2, the motion vector of a block falls within that area with a probability of 97.35.
Based on the results of Table (1) as we increase d there is a higher probability of getting the motion vector of the block inside the predicted search area.

It is also observed that to get the correct motion vector for video sequences such as Football, Susie and Foreman, which have fast and complex movements, d has to be higher than video sequences such as Clair that have slow movements. Increasing parameter "d" beyond 5 would cause no further improvement for most sequences, hence, those results are not included in Table (1).

**Table 1.** Probability of motion vector of a block falls with the rectangular area.

|  | d=1 | d=2 | d=3 | d=4 | d=5 |
|---|---|---|---|---|---|
| *Football* | 86.80 | 92.58 | 94.50 | 95.51 | 96.37 |
| *Claire* | 97.63 | 99.46 | 99.63 | 99.71 | 99.79 |
| *Susie* | 89.03 | 97.35 | 98.25 | 98.71 | 98.97 |
| *Garden* | 95.32 | 98.05 | 98.54 | 98.82 | 99.17 |
| *Trevor* | 92.58 | 99.06 | 99.46 | 99.62 | 99.69 |
| *Calendar* | 91.39 | 95.24 | 96.42 | 96.82 | 97.43 |
| *Stefan* | 95.28 | 98.05 | 98.48 | 98.67 | 98.98 |
| *Foreman* | 83.54 | 90.31 | 93.10 | 94.96 | 95.96 |
| ***Average*** | **91.45** | **96.26** | **97.30** | **97.85** | **98.30** |

The results of table 1 are based on the FS algorithm and the MAE criterion function. If in Equation (11) we use MSE instead of MAE then a new table is produced. According to our simulation results, for a specific d, the probability that the motion vector falls in a rectangular area is higher when MAE is used than when using MSE. For example, when using MAE function the probability that the motion vector falls within a 2×2 square, $\Pr(2) = 96.26\%$, while if MSE function is used then $\Pr(2) = 95.7\%$. Now let us investigate to see how d is effected when we change W. According to the simulation results for smaller W we can have smaller d in order to achieve same probability. For example, with W=15 we achieve $\Pr(3) = 97.3\%$ while W=31 results in $\Pr(5)$ of close to 97%. In our algorithm when d=5 and W=31, the average NSP is twice than when d is 3 and W is 15. This is in contrast to the FS algorithm that when W is changed from 15 to 31 the average NSP is quadrupled.

Now in the followings we present the details of our proposed search algorithm which is called *Predicted Vector Spectral Search Algorithm*, PVSSA.
1) Find the minimum and maximum values of the horizontal and vertical components of the five prediction motion vectors.
2) Extend the values of step 1 by d pixels to generate a search area.
3) Within the search area, the point that minimizes SAE is the matched block and the motion vector is computed for it.

The main criterion for designing this algorithm is its regularity and hence its simplicity in implementation. The search area of this algorithm is constant and unlike DS, 3SS, and 4SS extra search points are not added around the minimum point. A number of improvements can be suggested for the algorithm but at the expense of losing the regularity of the routine and hence those improvements are not discussed here.

The major difference between PSA and our proposed algorithm are:
1- The suggested algorithm uses temporal prediction vector. Since most objects in video sequence have inertial movement, the use of temporal correlation would increase the precision of the method.
2- PSA searches four regions. If overlapped points are not to be searched twice, extra information has to be kept and retrieved. The proposed algorithm searches each point only once, since it has an integrated search area.
3- The search area of the proposed algorithm is similar but very much smaller than that of the FS algorithm. Hence, same hardware implementations that are suggested for the FS can be applied to our algorithm. In short, the regularity of the suggested algorithm is its main advantage over PSA.

## 5 Simulation results

In our simulations maximum displacement (W) of 15, blocks of size 16*16 and images with Common Intermediate Format (CIF) are used. Our simulations were performed on MathWorks MATLAB, version 6.5.0, release 13. The hardware platform was a Pentium IV, 2.4GHz computer with 512 MB of RAM. Simulations used 30 frames of 8 standard video sequences. The suggested algorithm, with different values of d, is compared with DS, 4SS, 3SS and FS algorithms in terms of MSE, PSNR. For the blocks that are in the top row as well as the blocks in the left most and the right most columns not all of the five prediction vectors are available. Therefore, in place of any of the missing blocks (0, 0) motion vector is considered.

Table 2 shows average PSNR of the proposed algorithm for different values of d and compares them with that of FS algorithm. Taking d as 1 produces PSNRs that are close to those of FS except for Football, Susie and Foreman sequences. Increasing d to 2 would not get satisfactory results for the mentioned sequences. Expanding the search area with d equal to 3 generates outcomes that are close to FS in terms of PSNR. It is observed from Table 1 that the real motion vector is never predicted with 100 percent certainty. Therefore, if an error occurs in finding the motion vector of a block this error could propagate to other blocks and cause an accumulation of error.

Table 3 shows the average number of search points for a block for different window sizes. This number of search points is fixed for the FS algorithm and is 859. 45. It is interesting to notice that with one unit increase in d causes the average NSP to get almost doubled. It is further deduced that the motion vector of neighboring blocks in most sequences fall within a small area. For example, if the predicted motion vectors all point to one point then the rectangular are with d=3 covers 7x7 pixels. If the mentioned prediction vectors have a difference of 1 pixel then the search area is 8x8 pixels. This means that 64 points have to be searched. Values of Table 3 corresponding to d=3 are close to 64 indicating that generally the prediction motion vectors are very close to each other.

Table 4 contains the average MSE of each pixel of a frame. In this table, too, values corresponding to d=1, except for the *football, Susie* and the *foreman* sequences, are close to the FS algorithm.

**Table 2.** Comparing PSNR(dB) produced by test versions of the proposed algorithm and the FS algorithm.

|  | **FS** | **d = 1** | **d = 2** | **d = 3** | **d = 4** | **d = 5** |
|---|---|---|---|---|---|---|
| *Football* | 23.73 | 22.60 | 23.13 | 23.33 | 23.47 | 23.54 |
| *Claire* | 42.12 | 42.111 | 42.114 | 42.116 | 42.117 | 42.118 |
| *Susie* | 33.84 | 33.07 | 33.46 | 33.63 | 33.71 | 33.74 |
| *Garden* | 23.89 | 23.79 | 23.82 | 23.84 | 23.85 | 23.86 |
| *Trevor* | 33.59 | 33.49 | 33.54 | 33.55 | 33.56 | 33.57 |
| *Calendar* | 30.95 | 30.71 | 30.80 | 30.84 | 30.86 | 30.87 |
| *Stefan* | 24.33 | 24.19 | 24.25 | 24.26 | 24.27 | 24.28 |
| *Foreman* | 28.41 | 27.72 | 28.06 | 28.17 | 28.23 | 28.27 |
| ***Average*** | **30.108** | **29.710** | **29.897** | **29.967** | **30.008** | **30.031** |

**Table 3**. Average number of search points in PVSSA for different values of d.

|  | **d = 1** | **d = 2** | **d = 3** | **d = 4** | **d = 5** |
|---|---|---|---|---|---|
| *Football* | 44.14 | 79.83 | 121.79 | 169.07 | 221.28 |
| *Claire* | 15.70 | 35.69 | 62.49 | 96.14 | 136.57 |
| *Susie* | 32.36 | 60.59 | 95.07 | 136.16 | 183.27 |
| *Garden* | 14.77 | 34.19 | 61.13 | 95.69 | 137.33 |
| *Trevor* | 13.02 | 30.71 | 55.44 | 87.62 | 126.27 |
| *Calendar* | 13.88 | 34.87 | 63.76 | 102.18 | 147.02 |
| *Stefan* | 12.83 | 30.84 | 56.21 | 88.50 | 127.87 |
| *Foreman* | 31.44 | 62.54 | 100.43 | 146.16 | 192.76 |

**Table 4.** Comparing MSE produced by the FS algorithm and test versions of the suggested algorithm.

|  | **FS** | **d = 1** | **d = 2** | **d = 3** | **d = 4** | **d = 5** |
|---|---|---|---|---|---|---|
| *Football* | 275.84 | 359.13 | 317.46 | 302.82 | 293.74 | 288.77 |
| *Claire* | 4.205 | 4.211 | 4.209 | 4.207 | 4.206 | 4.206 |
| *Susie* | 27.09 | 32.62 | 29.66 | 28.51 | 27.97 | 27.74 |
| *Garden* | 259.68 | 274.68 | 272.37 | 271.21 | 270.66 | 269.98 |
| *Trevor* | 29.42 | 30.40 | 30.01 | 29.85 | 29.74 | 29.66 |
| *Calendar* | 53.22 | 56.23 | 55.06 | 54.60 | 54.32 | 54.23 |
| *Stefan* | 243.27 | 251.07 | 247.88 | 247.21 | 246.86 | 246.19 |
| *Foreman* | 95.01 | 120.94 | 108.95 | 105.44 | 103.59 | 102.55 |
| ***Average*** | **123.47** | **141.16** | **133.20** | **130.48** | **128.89** | **127.92** |

In Fig. 4 each graph shows the difference between PSNRs produced by the FS algorithm and those produced by PVSSA with different values of d. Parts a, b, c, and d of Fig. 4 respectively belong to 30 frames of football, Garden, Susie, and Calendar video sequences. It is apparent that there is not much variation for d being 3, 4, or 5. Also, for most sequences, results for increasing d above 5 are similar to those of 5.

Based on the graphs of Fig. 4 it is apparent that for the Football sequence due to the fast movements, using d=2 produces better results than d=1 and the difference between these results are more pronounced than in the other sequences. Furthermore, in all of the sequences the difference between d=1 and d=2 is more pronounced for the first frame and any other frame where there is sudden change in the scenery. For example, in Susie sequence frame 29 has different background than frame 28. Therefore, there is a larger difference in the mentioned graphs. When the movements are slow the temporal prediction vectors are more helpful in estimating the movements. But for the first frame or frames that the change in scenery is sudden the temporal prediction vector is of no use. This results in the apparent difference in the graphs of these type of frames.

It is deduced from Tables 2, 3, and 4 that the higher the value of d the more PSNR would result. This would increase the computational complexity, too. Therefore, considering the trade offs between the image quality and the complexity we base PVSSA on a fixed value of d equal to 3. Now let us define speedup ratio (SUR) parameter as in Equation (12). The SUR of proposed algorithm as compared to the FS algorithm is about 85.8% to 93.5%. This is equivalent to an increase in the search speed of 7 to 15.5 times that of the FS algorithm.

$$SUR = \frac{NSP_{FS} - NSP_{PVSSA}}{NSP_{FS}} \times 100\% \qquad (12)$$

In this equation $NSP_{FS}$ and $NSP_{PVSSA}$ are respectively the average NSP of the FS and PVSSA algorithms. Table 5 compares PSNR produced by PVSAA and other algorithms such as FS, PSA, 3SS, 4SS, and DS. It is seen that the proposed algorithm produces PSNRs that are closer to the results of FS, as compared to other algorithms.

Table 6 shows the average MSE resulted from application of PVSSA and some other algorithms. Again the proposed algorithm generates average MSE values that are closer to those of the FS algorithm.

Graphs of Fig. 5 show PSNR produced by PVSSA and five other algorithms by using 30 frames from 4 standard video sequences. In all cases the proposed algorithm performed better or at least performed as well as the other algorithms. The advantage of our algorithm is its regularity ease of implementation.

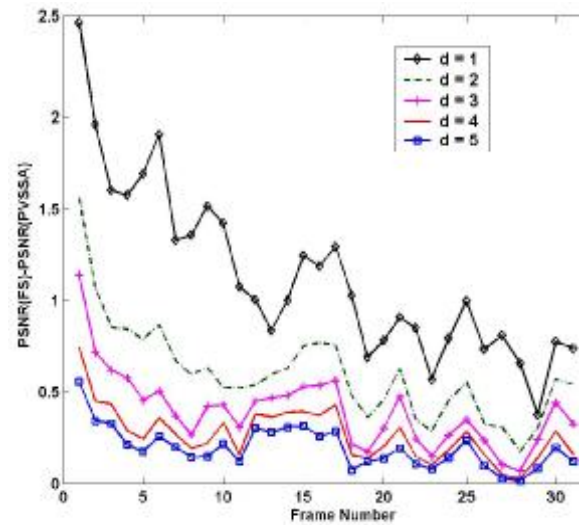

(a) football

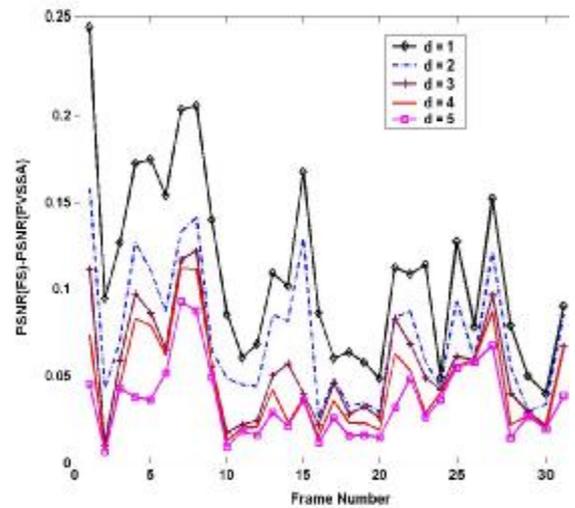

(b) Garden

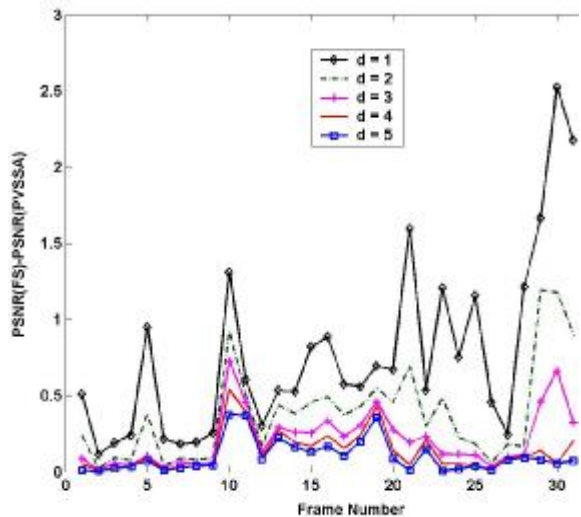

(c) Susie

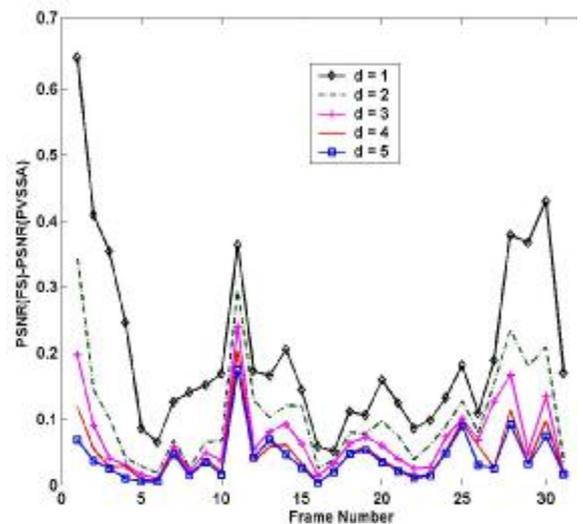

(d) Calendar

**Fig. 4.** Illustration of difference between PSNR of FS and suggested algorithm for varied *d* values.

**Table 5.** Comparison of different algorithms in terms of PSNR(dB)

|  | FS | 3SS | 4SS | DS | PSA | PVSSA(d=3) |
|---|---|---|---|---|---|---|
| *Football* | 23.73 | 22.63 | 22.66 | 22.84 | 22.37 | 23.33 |
| *Claire* | 42.12 | 42.00 | 42.03 | 42.11 | 42.06 | 42.116 |
| *Susie* | 33.84 | 32.64 | 32.82 | 33.43 | 32.71 | 33.63 |
| *Garden* | 23.89 | 21.27 | 23.33 | 23.74 | 23.78 | 23.84 |
| *Trevor* | 33.59 | 32.88 | 33.20 | 33.49 | 33.51 | 33.55 |
| *Calendar* | 30.95 | 28.42 | 30.44 | 30.78 | 30.70 | 30.84 |
| *Stefan* | 24.33 | 23.32 | 23.92 | 24.05 | 24.09 | 24.26 |
| *Foreman* | 28.41 | 27.48 | 27.79 | 28.00 | 27.77 | 28.17 |
| ***Average*** | **30.108** | **28.830** | **29.524** | **29.805** | **29.624** | **29.967** |

**Table 6.** Comparison of average MSE for different algorithms.

|  | FS | 3SS | 4SS | DS | PSA | PVSSA(d=3) |
|---|---|---|---|---|---|---|
| *Football* | 275.84 | 355.81 | 354.15 | 338.95 | 380.90 | 302.82 |
| *Claire* | 4.205 | 4.34 | 4.29 | 4.22 | 4.27 | 4.207 |
| *Susie* | 27.09 | 35.74 | 34.80 | 29.87 | 35.71 | 28.51 |
| *Garden* | 259.68 | 489.74 | 305.14 | 277.51 | 275.30 | 271.21 |
| *Trevor* | 29.42 | 34.78 | 32.22 | 30.33 | 30.29 | 29.85 |
| *Calendar* | 53.22 | 93.89 | 59.44 | 55.25 | 56.43 | 54.60 |
| *Stefan* | 243.27 | 316.02 | 270.40 | 261.29 | 257.07 | 247.21 |
| *Foreman* | 95.01 | 123.10 | 115.75 | 109.21 | 117.31 | 105.44 |
| ***Average*** | **123.47** | **181.678** | **147.024** | **138.329** | **144.660** | **130.48** |

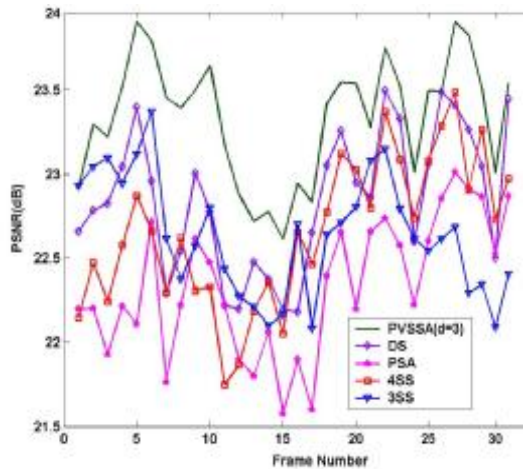

(a) football

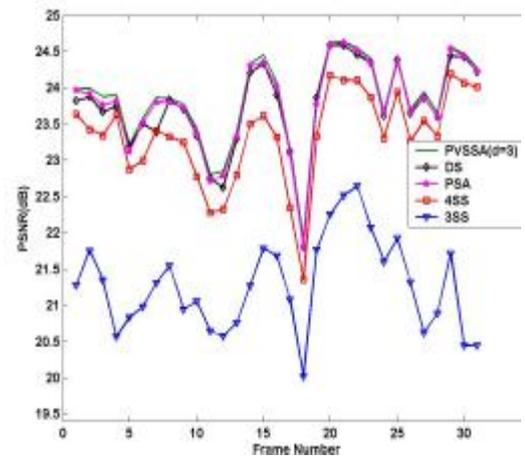

(b) Garden

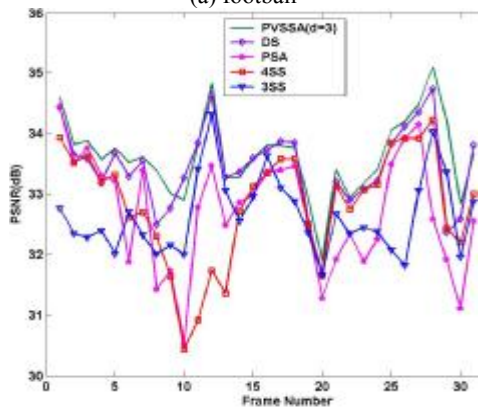

(c) Susie

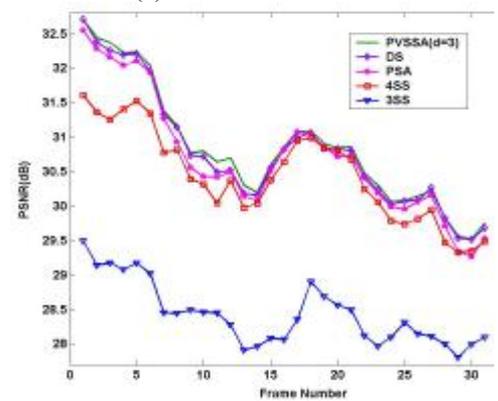

(d) Calendar

**Fig. 5.** Comparison of PSNR of traditional algorithms with *PVSSA(d=3)* for standard video sequences.

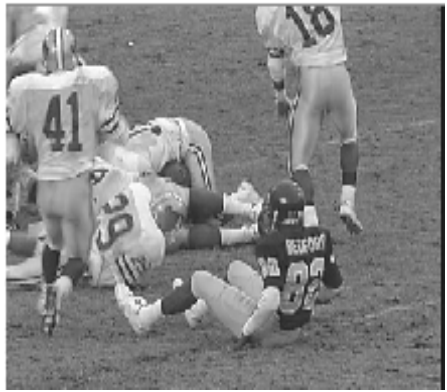

(a) original frame 20

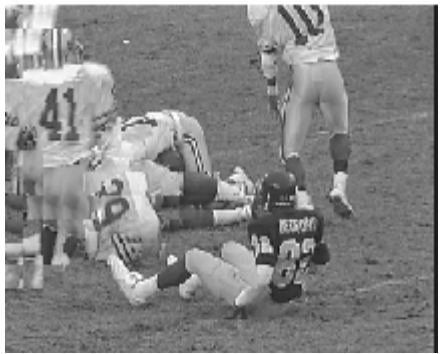

(b) FS algorithm

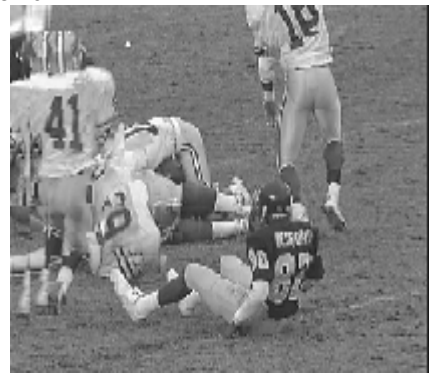

(c) PVSSA

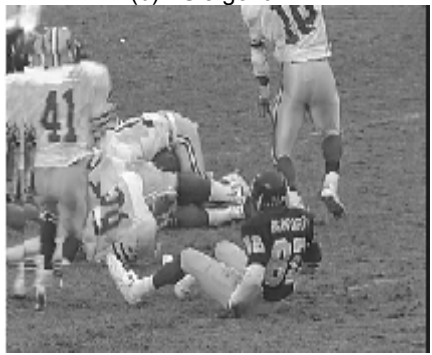

(d) PSA

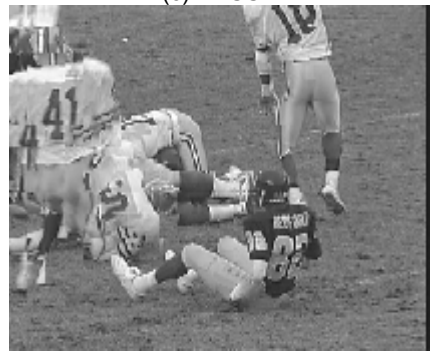

(e) DS algorithm

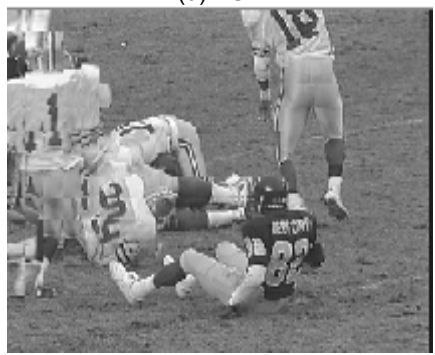

(f) 4SS algorithm

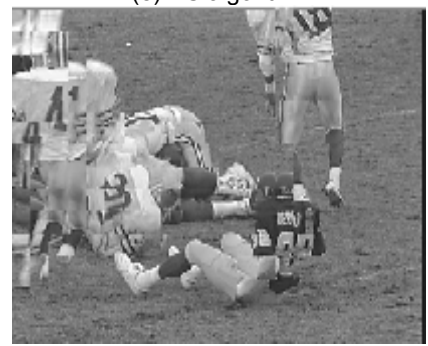

(g) 3SS algorithm

**Fig. 6.** Frame number 20 of football sequence3 reconstructed by different algorithms.

In Fig. 6(a) original 20[th] frame of football video sequence is shown. Fig. 6(b) through (e) are the reconstructed frames using frame number 19 of the sequence and the motion vectors generated with FS, PVSSA, PSA, DS, 4SS, and 3SS algorithms.

In Fig. 6(d), where PSA is used, digit 9 on the jersey of the player number 29 is not complete. Fig. 6(e), which DS algorithm is used to reconstruct it, besides digit 9 that is not clear, digit 1 on the jersey of player 41 is not complete either. In Fig. 6(f) 4SS algorithm is employed and digit 9 is not legible. Also, in the same figure digit 4 of player 41 is missing. Three step search algorithm in Fig. 6(g) has not been able to show digit 9, number 41, and number 82. The results of PVSSA and FS algorithms are very much the same. These are shown in Fig. 6(c) and (b).

## 6 Conclusions

In this paper an algorithm with regularity similar to that of FS was proposed. Taking advantage of spatial and temporal correlation among motion vectors of neighboring blocks the search area for each block was determined. This search area was initially determined by the spectrum that the motion vectors of the neighboring blocks covered. Then this area was further expanded by a predetermined value.

Based on the simulation results, for sequences with slow movements the MSE produced by the proposed algorithm is close to that of the FS algorithm. The computational complexity of the algorithm was very much smaller than the FS algorithm. 'Furthermore, the quality of the reconstructed images was in most cases superior to the results of many fast algorithms.

The same approach can be used for other applications [14-17].